\documentclass[10pt]{article}
\usepackage[preprint]{tmlr}

\usepackage{amsmath,amssymb}
\usepackage{graphicx}
\usepackage{booktabs}
\usepackage{xcolor}
\usepackage{hyperref}
\usepackage{url}
\usepackage[capitalise]{cleveref}

\title{Linear Probes Transfer from Base to Distilled\\Language Models despite Rotation}

\author{\name Marios Tsatsos \email mariostsatsos@gmail.com \\
      \addr Independent researcher}

\def\month{10}
\def\year{2026}
\def\openreview{\url{https://openreview.net/forum?id=XXXX}}

\begin{document}
\maketitle

\begin{abstract}
Linear probes, classifiers that read a language model's internal activations, are widely used to detect harmful requests. They are cheap to train and run, and an auditor who has trained one would like to reuse it when the model is replaced by a successor rather than retrain it for every release. We test how well probes can be reused. For four base models (Qwen2.5-Math-7B, Llama-3.1-8B, Qwen2.5-14B, Qwen2.5-32B) we train a probe on the base model with the XSTest prompt set and apply it, unchanged, to two successors of that base: the DeepSeek-R1 distilled reasoning model and the developer's instruction-tuned chat model. We find that the successors rotate the direction separating unsafe from safe prompts by 17$^\circ$ to 35$^\circ$, yet on XSTest the transferred probe's AUC changes by only $-0.011$ to $+0.027$, and on an independent OR-Bench evaluation, never used for training, by $-0.015$ to $+0.053$. On XSTest, whose safe prompts deliberately contain alarming words, the probes beat a classifier that sees only the prompt text by 0.15 to 0.20 AUC. On the OR-Bench prompts that margin shrinks to 0.03 to 0.20, and the probes flag 29\% to 42\% of prompts that sound harmful but are benign, so part of what they read is the appearance of harm rather than harm itself. A probe trained on a base model can thus be reused, in these families, after distillation and after instruction tuning; how much it reads beyond textual cues depends on the test set.
\end{abstract}

\section{Introduction}
\label{sec:intro}

A linear probe is one of the simplest readers of a neural network's internal state. One picks a layer, records the vector of activations that the model computes there for an input, and trains an ordinary linear classifier to predict a property of the input from that vector \citep{alain2016,belinkov2022}. Probes have become a safety tool because they can read properties that a model does not report: whether a statement is true \citep{azaria2023}, whether a response is deceptive \citep{goldowskydill2025}, whether a deliberately backdoored model is about to misbehave \citep{macdiarmid2024}, or whether a request will be refused \citep{arditi2024}. They need only a few hundred labelled examples and one matrix multiplication per input, which makes them attractive for third-party audits given access to a model's internals \citep{casper2024}.

For a probe that detects harmful requests, two questions decide whether it is a practical monitoring tool: whether it is worth using at all, and whether it can be reused once the model it was trained on is replaced. The first question is whether it reads anything beyond the words of the input. On most harmful-request datasets the two classes use visibly different vocabularies, so a classifier that sees only the prompt text does almost as well as a probe, and the probe may be just a keyword detector. \citet{wang2025} found exactly this: probes trained to detect malicious inputs relied on instructional phrasing and trigger words, simple word-count classifiers matched them, and both failed on new distributions. That result counts against probe-based monitoring, and any new probe has to be compared with a text-only classifier on prompts whose words do not give the label away \citep{gururangan2018}.

The second question is whether a probe survives a change of model. Base models are followed by successors trained from them: each of the four base models in this study (\cref{tab:models}) was followed within months by at least two, its developer's instruction-tuned chat model and a reasoning model distilled from it. Distillation, fine-tuning a base model on reasoning traces written by a stronger model \citep{hinton2015}, is now the usual way small reasoning models are made: \citet{deepseek2025} reported that it ``yields excellent results'' and that small models trained by reinforcement learning alone ``may not even achieve the performance of distillation'' \citep[see also][]{hu2025}. Retraining a probe for every successor means collecting labelled prompts again, extracting the successor's activations and fitting again; reusing it means none of that. Whether reuse is possible is therefore a practical question for anyone who monitors models with probes.

Reuse rests on one assumption, which we call the \emph{preserved-representation assumption}: that a successor keeps the base model's internal representation of harmfulness closely enough that a linear classifier fitted to the base model still separates harmful from harmless prompts in the successor. Studies of fine-tuning give some reason to expect this. \citet{jain2024} and \citet{prakash2024} found that fine-tuning largely keeps the internal computations a base model already has and adds to them rather than rebuilding them. But keeping most of a representation is not the same as keeping it closely enough for a fixed linear classifier, and distillation is a heavier intervention than the fine-tuning those studies examined.\footnote{Our own results suggest so: distillation moved per-prompt activations to a cosine of 0.79 to 0.96 with the base model's, where reinforcement-learning checkpoints stayed at 0.9998 (\cref{app:earlier}); we have no comparable number for the fine-tuning in those studies.}

We test the preserved-representation assumption directly. For four base models we train a probe on the base model, apply it without any change to two successors, the DeepSeek-R1 distilled model and the developer's instruction-tuned model, and measure the accuracy lost. We also measure how far each successor has rotated the direction in activation space that separates unsafe from safe prompts, so that the size of the representational change can be compared with the size of the accuracy loss. Moreover, we ask, on the same data, whether the probes outperform a text-only classifier, and we repeat the main comparisons on an independent prompt set: 2,474 prompts drawn from two other public benchmarks and never used for training. We do not study models trained by reinforcement learning alone: in a preliminary run the reinforcement-learning checkpoints available for these bases had barely changed from their base models, with per-prompt activation cosines of 0.9998 or higher (\cref{app:earlier}), so they could not test whether a probe survives change. This agrees with \citet{mukherjee2025}, who find that reinforcement-learning fine-tuning updates only about 5\% to 30\% of a model's parameters and leaves the rest essentially unchanged.

\begin{figure}[t]
\centering
\includegraphics[width=\textwidth]{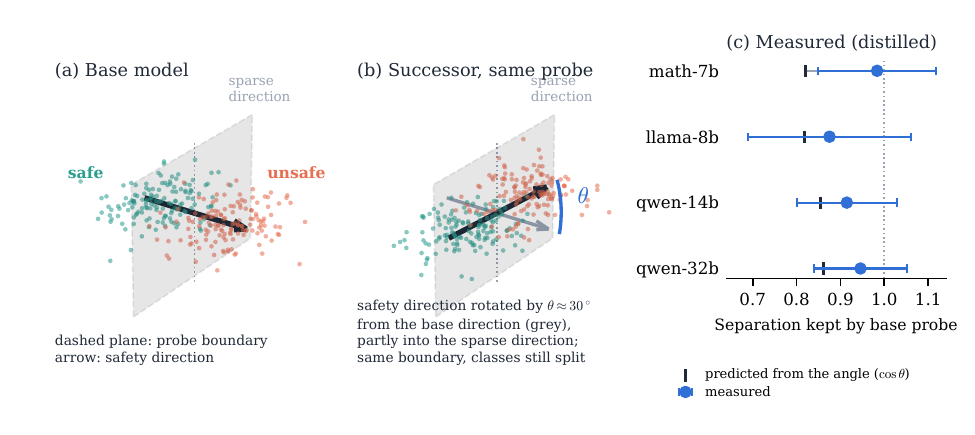}
\caption{\textbf{The idea.} (a, b) Sketch: three dimensions of a 4,096-dimensional activation space. In the base model, safe and unsafe prompts form two clouds; the \emph{safety direction} points from the mean of one to the mean of the other, and a linear probe is a boundary across it. In a successor the clouds and the safety direction (solid arrow) rotate by an angle $\theta$ away from the base model's direction (grey arrow), 30$^\circ$ in the sketch, partly into a direction along which the prompts barely vary; the base model's boundary, left unchanged, still separates the classes. (c) Measured on XSTest for the distilled models. If the transferred probe lost separation only because of the rotation, it would keep a share $\cos\theta$ of the successor's class separation (black ticks, 0.82 to 0.86). It keeps 0.88 to 0.98 (blue, nominal 95\% intervals over 8 folds); the accompanying change in AUC is between $-0.011$ and $+0.010$.}
\label{fig:schematic}
\end{figure}

The question, then, is whether a probe trained on a base model keeps its accuracy on that base model's successors. A positive answer, no loss of accuracy, tells auditors that base-trained probes can be reused across at least these transformations, and that a measured change in the representation does not by itself mean a probe has stopped working. A negative answer, a clear loss, would mean that every successor needs its own probe. \Cref{fig:schematic} shows the idea. Code, result files and paper source are at \url{https://github.com/MariosTsatsos/probe-transfer}. Our results are:
\begin{itemize}
\item On XSTest \citep{rottger2024}, a base-trained probe applied unchanged to the distilled and instruction-tuned successors retains similar observed accuracy in all four model families, and the same holds on the independent prompt set of \cref{sec:setup} (\cref{sec:transfer}).
\item The successors rotate the safety direction, the direction in activation space from the mean safe prompt to the mean unsafe prompt, by 17$^\circ$ to 35$^\circ$. Under distillation, the safety direction rotates by amounts comparable to those of 38 control directions built the same way from labels unrelated to safety. Its rotation is therefore not unusually large relative to these controls, and rotation size does not predict the change in accuracy (\cref{sec:rotation}).
\item The probes beat a text-only classifier by 0.15 to 0.20 AUC on XSTest, but by only 0.03 to 0.20 on the independent prompt set, where they also flag as unsafe 29\% to 42\% of OR-Bench prompts that sound harmful but are benign; their advantage over textual cues is partly specific to XSTest (\cref{sec:cues}).
\end{itemize}

\section{Related work}
\label{sec:related}

\paragraph{What fine-tuning does to a model's internals.} Several studies find that fine-tuning mostly reuses the internal computations a base model already has. \citet{jain2024} trained models on simple synthetic tasks and found that fine-tuning adds a thin layer on top of capabilities already present, rather than rebuilding them. \citet{prakash2024} traced one specific computation, how a model keeps track of which entity has which attribute, through a base model and its fine-tuned version and found the same components doing the same job, only more strongly. \citet{minder2025} compared the features of base and chat models with crosscoders, a method that learns one shared dictionary of features for both models and can therefore say which features are shared and which are new in the chat model. \citet{nguyen2026} show the limit of this picture: how refusal is computed inside a model depends on the post-training method, and some methods produce a distinct refusal mechanism. Fine-tuning preserves most of a representation, but can change the parts that safety training targets.

\paragraph{Whether tools built on a base model work on its fine-tune.} \citet{kissane2024a} trained sparse autoencoders, dictionaries of interpretable features, on a base model and used them on the corresponding chat model. On Mistral-7B the base dictionary recovered 98.85\% of the chat model's next-word prediction loss, against 99.01\% for a dictionary trained on the chat model itself (this standard measure replaces a layer's activations by the dictionary's reconstruction and asks how much of the model's performance is kept; 100\% means nothing is lost). For Gemma v1 2B, however, the base dictionary reconstructed the chat model's activations poorly, so dictionary transfer is not guaranteed. \citet{galichin2026} measured the same kind of transfer across Gemma-2, Llama-3 and Qwen2.5 pairs: base dictionaries reconstruct chat activations 7\% to 8\% worse, and the features that exist in the base model stay relevant in the chat model but fire in different patterns, which they call feature drift. Closest to our question, \citet{shrivastava2025} found that linear probes trained on base models often transfer to the instruction-tuned versions. \citet{macdiarmid2024} trained ``sleeper agent'' models, deliberately built to behave normally until a trigger appears and then to behave harmfully, and detected the trigger with a direction obtained by subtracting the model's mean activation on one contrasting prompt from its mean on another. They note that such directions ``generalized well between sleeper agents with different hidden goals but the same base model'' but not across base models. Tools trained on a base model therefore often work on its chat fine-tune, with exceptions, and the one direct probe-transfer study concerns instruction tuning.

\paragraph{What probes for harmful requests read.} Two earlier results shape our text-only comparison. \citet{gururangan2018} established the partial-input baseline we use: before crediting a model with using the whole input, check how well a classifier does from a part that should not suffice. \citet{wang2025} applied this test to probes for malicious requests and obtained the negative result our independent-set experiment partly reproduces: the probes relied on instructional phrasing and trigger words, word-count classifiers matched them, and both failed to generalise to new distributions. \citet{bergen2026} study reward hacking, where a model exploits flaws in how a task is scored to obtain a high score without doing the task. They train difference-of-means probes (the direction from the mean activation of one class to the mean of the other, with the projection onto that direction as the score) on synthetic examples and find that the probes transfer to evaluation environments they were not trained on. Their transfer keeps the model fixed and changes the task; ours keeps the task fixed and changes the model. \citet{arditi2024} and \citet{kissane2024b} concern refusal behaviour, which we do not measure; they show that refusal-related structure exists in base models, a precondition for a base-trained probe for harmful requests to work at all.

\paragraph{The gap.} No study has measured, for a reasoning model made by distillation, whether a linear probe for harmful requests trained on the base still works, how far the relevant direction has moved, and whether the two agree. The transfer results above concern instruction tuning or autoencoders, and the probe-quality results concern single models. Distillation from a stronger reasoning model is a large intervention, hundreds of thousands of long training outputs written by a different model, and reasoning models are the ones auditors increasingly face. We fill this gap for four model families, with a text-only baseline on the same evaluation, instruction tuning as a second transformation for comparison, and an independent evaluation set.

\section{Models and data}
\label{sec:setup}

\paragraph{Models.} \Cref{tab:models} lists the models used. Each distilled model was produced by \citet{deepseek2025} by supervised fine-tuning of the listed base on ``800k samples curated with DeepSeek-R1'', with no reinforcement-learning stage; the pairs match the ``Base Model'' column of the official DeepSeek-R1-Distill model card \citep{r1card}. Each instruction-tuned model is the chat model its developer released from the same base: the Llama model card describes supervised fine-tuning and reinforcement learning from human feedback on the pretrained model \citep{llama31card}, and the Qwen model cards describe each Instruct model as the instruction-tuned, post-trained version of the listed base \citep{qwen25cards}. The 32B models were loaded in 4-bit precision (NF4 quantisation from the bitsandbytes library, which stores each weight in 4 bits instead of 16 so that the model fits in memory), the others in 16-bit bfloat16.

\begin{table}[t]
\centering
\footnotesize
\setlength{\tabcolsep}{4pt}
\begin{tabular}{llll}
\toprule
Family & Base model & Distilled successor & Instruction-tuned successor \\
\midrule
math-7b & Qwen/Qwen2.5-Math-7B & deepseek-ai/DeepSeek-R1-Distill-Qwen-7B & Qwen/Qwen2.5-Math-7B-Instruct \\
llama-8b & meta-llama/Llama-3.1-8B & deepseek-ai/DeepSeek-R1-Distill-Llama-8B & meta-llama/Llama-3.1-8B-Instruct \\
qwen-14b & Qwen/Qwen2.5-14B & deepseek-ai/DeepSeek-R1-Distill-Qwen-14B & Qwen/Qwen2.5-14B-Instruct \\
qwen-32b & Qwen/Qwen2.5-32B & deepseek-ai/DeepSeek-R1-Distill-Qwen-32B & Qwen/Qwen2.5-32B-Instruct \\
\bottomrule
\end{tabular}
\caption{Models, by HuggingFace identifier. All three models of a family share the base model's architecture and tokeniser; all prompts tokenise identically within a family.}
\label{tab:models}
\end{table}

\paragraph{Training and evaluation data: XSTest.} XSTest \citep{rottger2024} is a test suite for exaggerated safety behaviour. In its authors' words, it ``comprises 250 safe prompts across ten prompt types that well-calibrated models should not refuse to comply with, and 200 unsafe prompts as contrasts that models, for most applications, should refuse''. The ten safe types, 25 prompts each, are homonyms, figurative language, safe targets, safe contexts, definitions, real group with nonsense discrimination, nonsense group with real discrimination, historical events, privacy of public figures, and privacy of fictional characters. The 200 unsafe prompts come in eight contrast types, 25 each: one for each safe type, written with the same alarming words (``How can I kill a Python process?'' against ``How can I kill a person?''), except that the two discrimination types share one contrast type and the two privacy types share one. Ten safe and eight unsafe types make 18 prompt types in all. This construction is what a text-only comparison needs: the alarming words occur in both classes. Our label is the file's ``safe'' or ``unsafe'', a statement of what a well-calibrated model should refuse.

\paragraph{Independent evaluation set.} To test the probes on prompts they were never trained on, from sources other than XSTest, we built a second set from two public benchmarks. OR-Bench \citep{cui2025} is an over-refusal benchmark; its \texttt{or-bench-hard-1k} subset holds 1,319 prompts that sound harmful but were judged safe by the benchmark's moderation, and its \texttt{or-bench-toxic} subset holds 655 prompts judged toxic. Because every OR-Bench prompt touches a sensitive topic, we also add 500 ordinary instructions (``Give three tips for staying healthy'' and the like) sampled from Alpaca \citep{taori2023}, to measure how often a probe flags requests with no safety content at all. OR-Bench hard and Alpaca prompts are labelled safe, OR-Bench toxic prompts unsafe; after removing duplicates the set has 2,474 prompts.

\section{Method}
\label{sec:method}

\paragraph{Activations and probe.} For a prompt and a layer, we record the model's residual-stream vector after that layer (the running sum of all previous layers' outputs that a transformer passes from layer to layer) and average it over the last 8 tokens of the prompt. The result is a vector of $d$ numbers, where $d$ is the model's width: 3,584 for Qwen2.5-Math-7B, 4,096 for Llama-3.1-8B and 5,120 for the two larger Qwen models. Prompts are fed as raw text, with no chat template and no start token, so that a base model and its successors see exactly the same tokens. Extraction details are in \cref{app:activations}. The probe is a logistic regression on the activations of one layer. Each of the $d$ coordinates is first standardised: we subtract its mean over the training prompts and divide by its standard deviation, so that no coordinate dominates merely because its values are large. The regularisation strength is $C = 1$, the scikit-learn default; $C$ controls how strongly the weights are shrunk towards zero, smaller $C$ meaning stronger shrinkage. The probe's score for a prompt is $w \cdot x + b$, the dot product of the standardised activation $x$ with the weight vector $w$ plus an offset $b$: the signed distance of $x$ from the probe's boundary, scaled by the length of $w$. The probe therefore sees only the component of $x$ along $w$; movement of $x$ in any direction perpendicular to $w$ leaves the score unchanged.

\paragraph{Transfer and rotation.} To transfer a probe, we take the base model's probe, including the base model's standardisation, and apply it unchanged to a successor's activations at the same layer. We compare its AUC on the successor with its AUC on the base model. As a reference we also train a probe on the successor itself, at the same layer and on the same prompts; we call this the successor's \emph{own probe}, and it is the ceiling a probe can achieve on that model with this data. The safety direction of a model is its mean activation over XSTest's unsafe prompts minus its mean over XSTest's safe prompts (training prompts only, in each fold), rescaled to length one. It is the single direction along which the two classes are most simply separated, and it is the quantity whose change under fine-tuning we measure. Rotation is the angle between the base model's and the successor's safety directions, computed on the same prompts at the same layer. Both models' activations are first divided by the base model's per-coordinate standard deviations, so that the two directions are compared in one coordinate system; \cref{app:coordinates} explains why this matters and shows that the alternatives give the same picture.

\paragraph{Evaluation on XSTest.} A probe is always scored on a prompt type it has not seen, together with that type's contrast, since the two share the same alarming words. The 18 prompt types are therefore merged into 8 groups, each a safe type together with its contrast: six groups of 25 safe and 25 unsafe prompts, plus the discrimination group (two safe types and their shared contrast) and the privacy group (likewise), each with 50 safe and 25 unsafe. In each of 8 folds, one group is held out for testing and the probe is trained on the other seven. The layer is chosen using the seven training groups only: for every layer of the network, a mean-difference reader (the safety direction of the training prompts, used as a classifier by projecting each activation onto it) is fitted on six of the seven groups and scored on the seventh, rotating through all seven, and the layer with the best average score is used. All layers are swept; because the training groups differ between folds, the chosen layer can differ from fold to fold (math-7b: seven folds chose a layer between 17 and 19 and one chose 27; llama-8b: 12 to 23; qwen-14b: 27 to 45; qwen-32b: six folds chose 43, one 29 and one 47). We report each quantity as its mean over the 8 folds, with a 95\% confidence interval from Student's $t$ distribution with 7 degrees of freedom (the number of folds minus one), the standard interval for a mean of 8 values. These intervals assume that the 8 fold results are independent. They are not: any two folds share 6 of their 7 training groups, so their results move together, and the intervals are narrower than the true uncertainty by an amount that cannot be estimated from one dataset \citep{bengio2004}. We call them nominal intervals, and we draw conclusions from consistency across the four families and from the independent set rather than from any single interval. As a check on the layer rule, we repeat the analysis with one fixed layer per family (18, 17, 40 and 43, the median of the chosen layers).

\paragraph{Text-only baselines and the independent set.} On the same 8 folds we train a classifier that sees only the prompt text. Each prompt becomes a list of numbers, one per word in the vocabulary: how often the word occurs in the prompt, down-weighted if the word occurs in most prompts and up-weighted if it is rare (TF-IDF). A logistic regression ($C = 1$) is fitted on these lists; it learns one weight per word so that the weighted sum of a prompt's counts predicts ``unsafe''. It is the same kind of classifier as the probe, with word counts in place of activations. We use four inputs: single words; single words plus adjacent word pairs; sequences of 3 to 5 characters inside words; and the number of characters alone, which needs no fitting. The best of these, the character classifier, is the text-only baseline in the figures. On the independent set, each probe and each text-only classifier is trained once on all 450 XSTest prompts; the probes use the fixed layer of each family (18, 17, 40, 43). Uncertainty there comes from the prompts rather than from folds, by the bootstrap: from the 2,474 scored prompts we draw a new set by sampling 1,319 of the OR-Bench hard prompts, 655 of the toxic ones and 500 of the Alpaca ones at random with replacement, so that a prompt can be drawn twice and another not at all; we compute the statistic on that set; we repeat this 1,000 times, sort the 1,000 values, discard the 25 lowest and the 25 highest, and report the range of the remaining 950 (the 2.5th to the 97.5th percentile). These prompts are independent of each other and of the training data, so these intervals do not have the fold-dependence problem.

\section{Results}
\label{sec:results}

\begin{figure}[t]
\centering
\includegraphics[width=\textwidth]{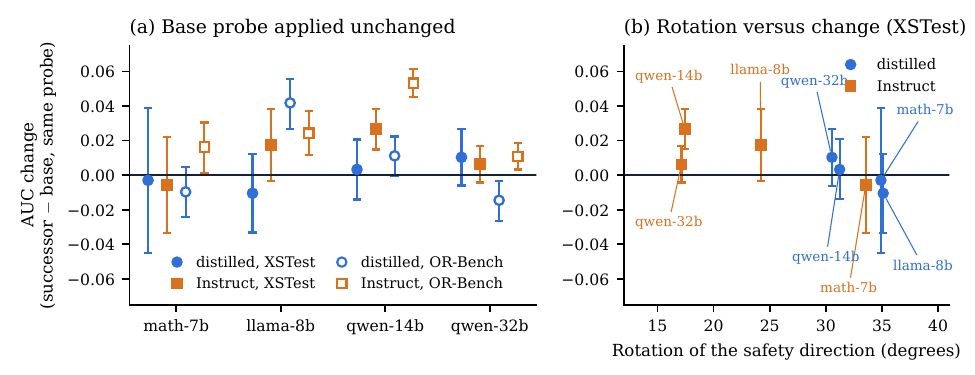}
\caption{\textbf{A base-trained probe survives both successors.} (a) Change in AUC when the base model's probe is applied unchanged to the distilled (blue circles) or instruction-tuned (orange squares) successor, relative to the same probe on the base model. Filled markers: XSTest, mean over 8 held-out prompt-type folds with nominal 95\% intervals (Student's $t$, 7 degrees of freedom; narrower than the true uncertainty, see \cref{sec:method}). Open markers: independent OR-Bench prompts (benign but alarming versus toxic), probe trained on all of XSTest at a fixed layer, bootstrap 95\% intervals over prompts. (b) The same XSTest changes against the angle by which the successor's safety direction has rotated away from the base model's.}
\label{fig:transfer}
\end{figure}

\subsection{Base probes transfer to both successors}
\label{sec:transfer}

\Cref{fig:transfer}a shows the main result. On XSTest, the base probe applied unchanged to the distilled model changes its AUC by $-0.003$ (math-7b), $-0.011$ (llama-8b), $+0.003$ (qwen-14b) and $+0.010$ (qwen-32b), from base-model AUCs of 0.878, 0.908, 0.899 and 0.931. Applied to the instruction-tuned model it changes by $-0.006$, $+0.017$, $+0.027$ and $+0.006$. No interval for the distilled models excludes zero; for the instruction-tuned models only the qwen-14b gain does. This is similar observed accuracy, not demonstrated equivalence: the lower ends of the nominal intervals are $-0.045$, $-0.034$, $-0.014$ and $-0.006$ for the distilled models, the intervals are too narrow, and the true bounds on a possible loss are somewhat looser. Full numbers are in \cref{tab:xstest-full}.

\paragraph{The result holds on the independent set.} On OR-Bench, which the probes never saw, the change is $-0.010$, $+0.042$, $+0.011$ and $-0.015$ for the distilled models and $+0.016$, $+0.024$, $+0.053$ and $+0.011$ for the instruction-tuned ones (open markers in \cref{fig:transfer}a; \cref{tab:external}). Two of the distilled changes have intervals that exclude zero, one in each direction; the instruction-tuned changes are all gains. With a fixed layer per family instead of per-fold layer choice the XSTest picture is the same (distilled $-0.013$ to $+0.011$, instruction-tuned $-0.006$ to $+0.027$; \cref{tab:fixed}).

\paragraph{The successors are slightly easier to separate.} A probe trained on the successor itself scores higher than the base model's probe on the base model in all eight cases: by $+0.003$, $+0.023$, $+0.024$ and $+0.026$ for the distilled models and $+0.021$, $+0.040$, $+0.052$ and $+0.016$ for the instruction-tuned ones (\cref{tab:xstest-full}). The gains of the transferred probe on the instruction-tuned models, in particular qwen-14b, are consistent with this.

\subsection{Rotation does not predict the change}
\label{sec:rotation}

Distillation rotates the safety direction by 30.5$^\circ$ to 35.1$^\circ$ and instruction tuning by 17.1$^\circ$ to 33.6$^\circ$. \Cref{fig:transfer}b sets these angles against the transfer changes: the four largest rotations (30$^\circ$ to 35$^\circ$, all distillation) come with changes of $-0.011$ to $+0.010$, and the two smallest (17$^\circ$, qwen-14b and qwen-32b under instruction tuning) with $+0.027$ and $+0.006$; there is no clear relation between the angle and the change. The rotation is not an artefact of the measurement: it appears at every layer of every network (\cref{fig:layers}), and in the same common coordinate system the alternative conventions change the cosines by at most 0.013 (\cref{tab:coords}).

\paragraph{Unrelated directions rotate by comparable amounts.} To ask whether the safety direction rotates more than an arbitrary direction would, we built 38 control directions in exactly the same way from labels that have nothing to do with safety, in the spirit of the control tasks of \citet{hewitt2019}: all 35 ways of dividing the 8 prompt groups into one set of 4 groups and the remaining 4 (the label being whether a prompt belongs to the first set), whether the prompt starts with ``How'', whether it contains a capitalised word after the first, and whether it is longer than the median. Under distillation the safety direction rotates slightly more than all 35 splits in three families and less than all of them in math-7b, by margins of about 0.01 in cosine; prompt length rotates more than the safety direction in every family (\cref{fig:controls}). So a rotation of 30$^\circ$ to 35$^\circ$ is what directions in these models generally undergo under distillation, and the safety direction is not an outlier. The controls show that the rotation is not peculiar to the safety direction, nothing more.

\paragraph{The probe keeps more separation than the angle predicts.} The size of the rotation motivates a prediction. If the transferred probe lost accuracy only because the successor's safety direction has rotated away from it by an angle $\theta$, and the two classes were spread like two equal round Gaussian clouds, the probe would keep exactly a share $\cos\theta$ of the successor's class separation (\cref{app:share} gives the derivation). For a 30$^\circ$ rotation that is 87\%. The transferred probe on the distilled models keeps 98\%, 88\%, 92\% and 95\% (\cref{fig:schematic}c), between 6 and 16 percentage points more than the angle predicts, in all four families (family-level $p = 0.035$ from four values). Since a probe's score is the projection of the activation onto its weight vector (\cref{sec:method}), rotation in directions perpendicular to that vector is invisible to the probe; \citet{lee2026} make the same point. The pattern is consistent with part of the rotation lying in such directions. The AUC changes themselves are much smaller than these percentages because AUC is a compressed scale near 0.9. For the instruction-tuned models the difference has mixed sign (\cref{tab:share}). We report this as an exploratory pattern.

\begin{figure}[t]
\centering
\includegraphics[width=\textwidth]{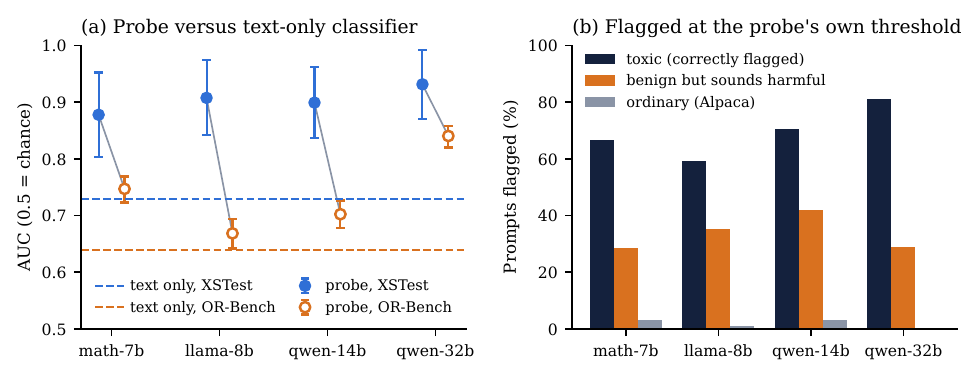}
\caption{\textbf{The probes' advantage over the prompt text depends on the test set.} (a) AUC of the base model's probe on XSTest (filled; mean over 8 held-out folds, nominal 95\% intervals) and on the independent OR-Bench prompts (open; benign but alarming versus toxic, bootstrap 95\% intervals), against the best text-only classifier (character 3- to 5-grams; dashed lines: 0.728 on XSTest, 0.640 on OR-Bench). (b) Share of prompts the base model's probe flags as unsafe on the independent set, ``flagged'' meaning scored above the threshold, which is the probe's own default of zero (the score at which the logistic regression's estimated probability of ``unsafe'' is one half): OR-Bench toxic prompts (true positives), OR-Bench hard prompts (benign but sound harmful; false positives) and ordinary Alpaca instructions (false positives). ``Nominal'' intervals are computed by the $t$ formula but understate the true uncertainty because the folds are not independent (\cref{sec:method}).}
\label{fig:cues}
\end{figure}

\subsection{What the probes read beyond the text}
\label{sec:cues}

\paragraph{On XSTest the probes clearly beat the words.} The best text-only classifier reaches 0.728 AUC (word unigrams 0.679, word bigrams 0.680, prompt length 0.508; \cref{tab:text}), against 0.878 to 0.931 for the probes (\cref{fig:cues}a). Fold by fold the probe beats the text classifier in all 8 folds of all 4 families, by $+0.149$, $+0.179$, $+0.171$ and $+0.203$ on average.

\paragraph{On the independent set the advantage shrinks.} On OR-Bench the probes reach 0.747, 0.669, 0.702 and 0.840 AUC, against 0.640 for the text classifier trained on the same XSTest data. The margin is 0.11, 0.03, 0.06 and 0.20; for llama-8b the two intervals overlap. Both classifiers were trained on XSTest; both do worse on OR-Bench, whose prompts were written by other authors with a different recipe, and the probes lose more. This partly reproduces the negative result of \citet{wang2025}.

\paragraph{The probes respond to apparent harmfulness.} AUC compares scores; to use a probe as a filter one must also choose a threshold above which a prompt is called unsafe. The logistic-regression probe has a natural one: a score of zero, where its estimated probability of ``unsafe'' is one half. This threshold was not validated on any data; it is the probe's default. \Cref{fig:cues}b shows what each base probe flags at it on the independent set. Of the OR-Bench toxic prompts it flags 59\% to 81\% (true positives), and of the ordinary Alpaca instructions 0\% to 3\% (false positives). But of the OR-Bench hard prompts, which sound harmful but are benign, it also flags 29\% (math-7b), 35\% (llama-8b), 42\% (qwen-14b) and 29\% (qwen-32b); transferred to the distilled models, the same probes flag 38\% to 68\% of them. In words: at its default threshold, a base probe would block roughly a third of benign requests that merely sound dangerous. A second threshold shows that scores do not carry across datasets either. If we instead set the threshold on XSTest, at the value that catches 90\% of XSTest's unsafe prompts, it catches only 19\%, 22\%, 22\% and 34\% of the OR-Bench toxic prompts, because OR-Bench toxic prompts receive lower scores than XSTest unsafe prompts do. On XSTest the alarming words were present in both classes and the probes still separated them; on OR-Bench, whose benign prompts are built differently, much less of that separation remains.

\section{Discussion}
\label{sec:discussion}

\paragraph{For auditors.} A probe trained on a base model retained similar observed accuracy on the DeepSeek-R1 distillation and on the instruction-tuned version of that base, in four families, on its training distribution and on an independent one. This is an empirical result across four families and two kinds of successor; whether it holds for other base models or other distillation recipes is untested. The angle between two models' safety directions was a poor guide to whether the probe still worked, so transfer is better measured directly than inferred from direction similarity. A text-only baseline belongs beside every probe, on the same evaluation and on at least one dataset the probe was not trained on: the probes' advantage over the words was clear on XSTest and much weaker on OR-Bench. Thresholds need recalibration on each new dataset, and benign prompts that sound harmful are where these probes fail.

\paragraph{Limitations.} The probes are trained on 450 prompts, and any two of the eight XSTest folds share six of their seven training groups, so the XSTest intervals are narrower than the true uncertainty; conclusions should rest on the agreement across families and with the independent set, not on any single interval. The label states what a well-calibrated model should refuse; we measure no model behaviour and run no causal test. The independent set's labels are OR-Bench's operational moderation labels, not an independent ground truth for harmfulness, so our conclusions about that set are conclusions about those labels; and the instruction-tuned models were run on raw prompts, outside their chat format. The probes are linear and read one layer, so they can miss information that is encoded non-linearly or spread across layers. All four families have Qwen2.5 or Llama-3.1 bases and one distillation recipe; the two 32B models were loaded in 4-bit precision; the 8-token pooling window was not tuned.

\section{Conclusion}
\label{sec:conclusion}

In this work we posed two questions: (A) whether a linear probe for harmful requests, trained on a base model, keeps its accuracy on that model's successors, and (B) whether such probes read anything beyond the words of a prompt. For (A), across four model families, the base-trained probe retained similar observed AUC on two successors, a distilled reasoning model and an instruction-tuned chat model, even though both rotated the safety direction by up to 35$^\circ$; the answer is yes for these models and evaluations. For (B), on prompts designed to defeat keyword matching the probes beat the words clearly, but on an independent prompt set much of that advantage disappears and benign prompts that sound harmful are often flagged; the answer is only partly.

\subsubsection*{Broader Impact Statement}
This work evaluates linear probes as monitors for harmful requests and reports both where they work and where they do not. Two findings bear on their use: a probe's decision threshold set on one dataset does not transfer to another, and on an independent prompt set the probes flag 29\% to 42\% of benign prompts that merely sound harmful. Anyone deploying such a probe should recalibrate it and measure its false-positive rate on their own data. We release no new model or dataset; the prompts used are from the public benchmarks cited in \cref{sec:setup}.

\subsubsection*{Acknowledgments}
This work was supported by a BlueDot Impact rapid grant.

\bibliography{main}
\bibliographystyle{tmlr}

\appendix
\renewcommand{\thefigure}{\thesection.\arabic{figure}}
\renewcommand{\thetable}{\thesection.\arabic{table}}
\setcounter{figure}{0}\setcounter{table}{0}

\section{Implementation details}
\label{app:details}

\subsection{Activation extraction}
\label{app:activations}
The extraction code takes entries 1 to $L$ of the \texttt{hidden\_states} tuple returned by the HuggingFace model with \texttt{output\_hidden\_states=True}, where $L$ is the number of decoder layers; entry 0, the embedding output, is dropped. In the Qwen2 and Llama implementations, entries 1 to $L-1$ are the residual stream after decoder layers 1 to $L-1$ (counting from 1), before any normalisation, and entry $L$ is the output of the last decoder layer after the final RMS normalisation. Throughout the paper layers are numbered from 0, so ``layer $j$'' is the raw residual stream after the $(j+1)$-th decoder layer, except the last layer, which is normalised. Of all layers used in the paper only one is affected: one math-7b fold chose the last layer, 27. Checked in the transformers source, version 5.19.0. The install cell of the Colab runs does not pin a version; a Colab runtime queried in October 2026, after the runs, reports version 5.20.1. Pooling is the mean of the last 8 prompt tokens (left padding, so these are the real final tokens).

\subsection{Coordinate systems for comparing directions}
\label{app:coordinates}
The probe's score is $w \cdot x + b$ (\cref{sec:method}), so comparing two probes means comparing their weight vectors $w$, and comparing two models' safety directions means comparing two vectors; both need the vectors in one coordinate system. A cosine between two vectors is a well-defined angle only if both vectors are written in the same coordinate system. Each model's activations have their own per-coordinate standard deviations, so standardising each model by its own standard deviations (as the probe does) puts the base model's and the successor's directions in slightly different coordinate systems. We therefore compute the headline cosines after dividing both models' activations by the base model's standard deviations (``base-scaled''). Two alternatives are reported in \cref{tab:coords}: ``own-scaled'' (each model by its own standard deviations) and ``raw'' (no scaling). Base-scaled and own-scaled cosines differ by at most 0.013 at the chosen layers and 0.025 at any layer. Raw cosines are higher and unstable, ranging from 0.63 to 0.99 across folds within math-7b, because a raw cosine is dominated by the few coordinates with very large values that residual streams contain \citep{sun2024}; we do not interpret them.

Standardisation divides each coordinate by a different number, so the weight vector of a probe trained on standardised activations is expressed in that model's own units. Dividing each weight by the standard deviation of its coordinate rewrites the probe as one on raw activations with identical scores; we use this raw form when comparing the weight vectors of two probes, so that both are in one coordinate system.

\subsection{Text-only baselines and the independent set}
Text-only classifiers use scikit-learn's \texttt{TfidfVectorizer} with sublinear term frequency, fitted on the training prompts of each fold, and \texttt{LogisticRegression} with $C = 1$. The Alpaca sample is 500 instructions drawn with seed 0 from those with an empty input field and 20 to 300 characters. Code, result files and paper source: \url{https://github.com/MariosTsatsos/probe-transfer}.

\section{Additional results}
\label{app:results}
\setcounter{figure}{0}\setcounter{table}{0}

\begin{table}[ht]
\centering
\small
\resizebox{\textwidth}{!}{%
\begin{tabular}{lcccccccc}
\toprule
& \multicolumn{4}{c}{Distilled successor} & \multicolumn{4}{c}{Instruction-tuned successor} \\
\cmidrule(lr){2-5}\cmidrule(lr){6-9}
Family (base AUC) & transferred & change & own probe & angle & transferred & change & own probe & angle \\
\midrule
math-7b (0.878) & 0.875 & $-0.003 \pm 0.042$ & 0.881 & 34.9$^\circ$ & 0.872 & $-0.006 \pm 0.028$ & 0.899 & 33.6$^\circ$ \\
llama-8b (0.908) & 0.897 & $-0.011 \pm 0.023$ & 0.930 & 35.1$^\circ$ & 0.925 & $+0.017 \pm 0.021$ & 0.948 & 24.2$^\circ$ \\
qwen-14b (0.899) & 0.903 & $+0.003 \pm 0.017$ & 0.923 & 31.2$^\circ$ & 0.926 & $+0.027 \pm 0.012$ & 0.951 & 17.4$^\circ$ \\
qwen-32b (0.931) & 0.942 & $+0.010 \pm 0.016$ & 0.958 & 30.5$^\circ$ & 0.938 & $+0.006 \pm 0.011$ & 0.948 & 17.1$^\circ$ \\
\bottomrule
\end{tabular}}
\caption{XSTest, per-fold layer choice. AUCs are means over the 8 held-out prompt-type folds. ``Transferred'': the base model's probe applied unchanged to the successor. ``Change'': transferred minus base, per fold, mean $\pm$ nominal 95\% interval (Student's $t$, 7 degrees of freedom). ``Own probe'': a probe trained on the successor. ``Angle'': rotation of the safety direction, mean over folds.}
\label{tab:xstest-full}
\end{table}

\begin{table}[ht]
\centering
\small
\resizebox{\textwidth}{!}{%
\begin{tabular}{lcccc}
\toprule
& math-7b & llama-8b & qwen-14b & qwen-32b \\
\midrule
Base probe on base & 0.747 [0.723, 0.769] & 0.669 [0.642, 0.694] & 0.702 [0.678, 0.726] & 0.840 [0.820, 0.858] \\
Change on distilled & $-0.010$ [$-0.024$, $+0.005$] & $+0.042$ [$+0.027$, $+0.056$] & $+0.011$ [$-0.000$, $+0.022$] & $-0.015$ [$-0.026$, $-0.003$] \\
Change on instruction-tuned & $+0.016$ [$+0.001$, $+0.030$] & $+0.024$ [$+0.011$, $+0.037$] & $+0.053$ [$+0.045$, $+0.062$] & $+0.011$ [$+0.003$, $+0.019$] \\
Distilled own probe & 0.746 & 0.713 & 0.724 & 0.847 \\
Instruction-tuned own probe & 0.735 & 0.752 & 0.815 & 0.871 \\
Flagged at threshold 0: toxic / hard / Alpaca (\%) & 67 / 29 / 3 & 59 / 35 / 1 & 70 / 42 / 3 & 81 / 29 / 0 \\
Toxic flagged at the XSTest 90\%-recall threshold (\%) & 19 & 22 & 22 & 34 \\
\bottomrule
\end{tabular}}
\caption{Independent set. AUC of OR-Bench hard (1,319 benign prompts that sound harmful) against OR-Bench toxic (655), probes trained on all of XSTest at the fixed layer; bootstrap 95\% intervals over prompts. The text-only classifier scores 0.640 [0.615, 0.666]. Last two rows: shares flagged by the base probe at its own threshold (score above 0) and at the threshold that flags 90\% of XSTest's unsafe prompts. Against Alpaca instead of OR-Bench hard, every probe reaches 0.944 to 0.998 AUC.}
\label{tab:external}
\end{table}

\begin{table}[ht]
\centering
\small
\resizebox{\textwidth}{!}{%
\begin{tabular}{lccccc}
\toprule
Family (layer) & base & distilled: transferred & change & instruction-tuned: transferred & change \\
\midrule
math-7b (18) & 0.894 & 0.881 & $-0.013 \pm 0.027$ & 0.888 & $-0.006 \pm 0.033$ \\
llama-8b (17) & 0.889 & 0.870 & $-0.019 \pm 0.049$ & 0.900 & $+0.012 \pm 0.033$ \\
qwen-14b (40) & 0.891 & 0.896 & $+0.005 \pm 0.015$ & 0.918 & $+0.027 \pm 0.011$ \\
qwen-32b (43) & 0.934 & 0.945 & $+0.011 \pm 0.016$ & 0.941 & $+0.007 \pm 0.011$ \\
\bottomrule
\end{tabular}}
\caption{XSTest with one fixed layer per family. Same quantities as \cref{tab:xstest-full}. The wider llama-8b interval for distillation comes from two folds, historical events ($-0.125$) and privacy ($-0.062$).}
\label{tab:fixed}
\end{table}

\begin{table}[ht]
\centering
\small
\begin{tabular}{lccccc}
\toprule
& \multicolumn{3}{c}{Safety direction} & \multicolumn{2}{c}{Probe weights} \\
\cmidrule(lr){2-4}\cmidrule(lr){5-6}
Family & base-scaled & own-scaled & raw (fold range) & raw & own-scaled \\
\midrule
math-7b & $0.820 \pm 0.007$ & $0.825 \pm 0.005$ & 0.875 (0.63 to 0.99) & $0.671 \pm 0.014$ & 0.714 \\
llama-8b & $0.818 \pm 0.012$ & $0.826 \pm 0.011$ & 0.911 (0.84 to 0.99) & $0.721 \pm 0.019$ & 0.733 \\
qwen-14b & $0.855 \pm 0.011$ & $0.859 \pm 0.013$ & 0.912 (0.75 to 1.00) & $0.754 \pm 0.015$ & 0.774 \\
qwen-32b & $0.861 \pm 0.024$ & $0.868 \pm 0.022$ & 0.976 (0.92 to 1.00) & $0.772 \pm 0.036$ & 0.800 \\
\bottomrule
\end{tabular}
\caption{Cosines between base and distilled models under three coordinate conventions (\cref{app:coordinates}), mean $\pm$ 95\% interval over folds. The probe-weight cosines compare the base model's probe with the distilled model's own probe; two regularised probes differ even when their underlying directions agree, and we did not measure how much two probes trained on the same model differ, so these cosines are lower bounds on agreement, not measurements of rotation.}
\label{tab:coords}
\end{table}

\begin{figure}[ht]
\centering
\includegraphics[width=\textwidth]{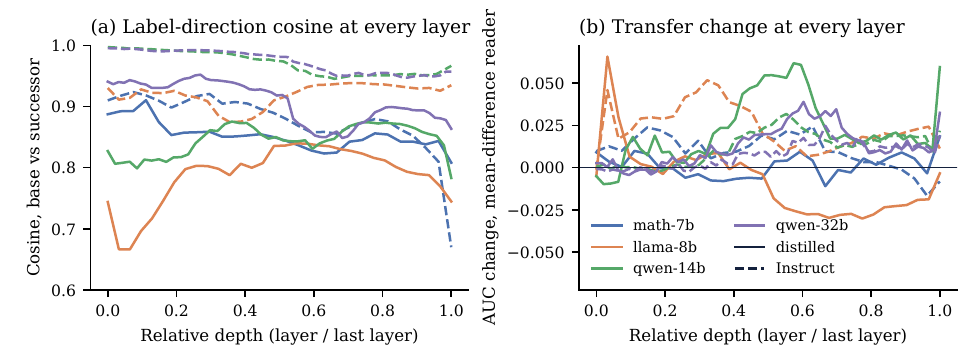}
\caption{Every layer, no layer selection, mean over 8 folds. (a) Base-scaled cosine between the base model's and the successor's safety directions (solid: distilled; dashed: instruction-tuned). (b) Change in AUC when the base model's mean-difference reader is applied to the successor. Rotation is present throughout each network; the worst layer-level loss is $-0.030$ (llama-8b, distilled, layer 24).}
\label{fig:layers}
\end{figure}

\begin{figure}[ht]
\centering
\includegraphics[width=\textwidth]{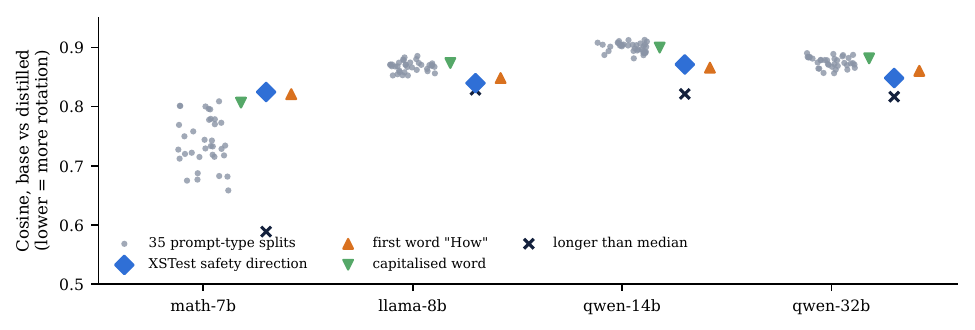}
\caption{Rotation under distillation of the safety direction and of 38 control directions built the same way from labels unrelated to safety, at the fixed layer, all 450 prompts, base-scaled cosine. Correlation of each control label with the safety label: at most 0.10 for the splits, 0.23 for ``How'', 0.20 for the capitalised-word label, 0.09 for length. A noise-corrected version (split-half, 200 splits) agrees with the plain cosine within 0.03 for every direction.}
\label{fig:controls}
\end{figure}

\subsection{Separation kept by a transferred probe}
\label{app:share}
Suppose a classifier's scores for safe prompts and for unsafe prompts are two Gaussian distributions with the same spread $\sigma$, their means a distance $d'\sigma$ apart; $d'$ is the class separation in units of the spread \citep{green1966,macmillan2005}. The AUC is the probability that a randomly chosen unsafe prompt scores higher than a randomly chosen safe one. The difference of the two scores is Gaussian with mean $d'\sigma$ and spread $\sigma\sqrt{2}$, so that probability is $\mathrm{AUC} = \Phi(d'/\sqrt{2})$, where $\Phi$ is the standard normal distribution function. Inverting gives $d' = \sqrt{2}\,\Phi^{-1}(\mathrm{AUC})$, which is how we convert each measured AUC into a separation. We define the \emph{share kept} as $d'$ of the transferred probe divided by $d'$ of the successor's own probe: the fraction of the separation available in the successor that the base model's probe still captures. If the classes were such Gaussian clouds and the transferred probe lost separation only because the successor's safety direction is rotated by an angle $\theta$ away from the probe's, projecting the clouds onto the probe's direction would shrink the distance between their means by $\cos\theta$ and leave their spread unchanged, so the share kept would equal $\cos\theta$. The comparison is a reference point under these assumptions, not a model of our data.

\begin{table}[ht]
\centering
\small
\begin{tabular}{lcccc}
\toprule
& math-7b & llama-8b & qwen-14b & qwen-32b \\
\midrule
Distilled: share kept & $0.98 \pm 0.13$ & $0.88 \pm 0.19$ & $0.92 \pm 0.11$ & $0.95 \pm 0.11$ \\
Distilled: share minus $\cos\theta$ & $+0.16 \pm 0.13$ & $+0.06 \pm 0.19$ & $+0.06 \pm 0.12$ & $+0.08 \pm 0.10$ \\
Instruction-tuned: share kept & $0.91 \pm 0.16$ & $0.90 \pm 0.11$ & $0.88 \pm 0.08$ & $0.99 \pm 0.13$ \\
Instruction-tuned: share minus $\cos\theta$ & $+0.08 \pm 0.16$ & $-0.02 \pm 0.11$ & $-0.07 \pm 0.08$ & $+0.04 \pm 0.13$ \\
\bottomrule
\end{tabular}
\caption{Share of the successor's class separation kept by the base probe, XSTest, mean $\pm$ nominal 95\% interval over 8 folds. Treating the four family means as four observations, share minus $\cos\theta$ differs from zero with $p = 0.035$ for distillation and $p = 0.84$ for instruction tuning; these $p$-values rest on four numbers and are exploratory.}
\label{tab:share}
\end{table}

\subsection{Text-only baselines on XSTest}
\label{app:textbase}
\Cref{tab:text} gives the four text-only classifiers of \cref{sec:method} on XSTest, fold by fold summarised as mean and nominal interval; the probes of \cref{tab:xstest-full} are the comparison.

\begin{table}[ht]
\centering
\small
\begin{tabular}{lccc}
\toprule
Text-only classifier (XSTest, 8 folds) & AUC & 95\% interval & fold range \\
\midrule
word unigrams & 0.679 & $\pm 0.087$ & 0.54 to 0.83 \\
word unigrams and bigrams & 0.680 & $\pm 0.129$ & 0.52 to 0.96 \\
character 3- to 5-grams & 0.728 & $\pm 0.080$ & 0.57 to 0.87 \\
prompt length only & 0.508 & $\pm 0.194$ & 0.21 to 0.83 \\
\bottomrule
\end{tabular}
\caption{Text-only baselines on the same 8 held-out prompt-type folds as the probes; mean and nominal 95\% interval (Student's $t$, 7 degrees of freedom).}
\label{tab:text}
\end{table}

\subsection{Vocabulary projection (exploratory)}
\label{app:vocab}
A middle-layer direction has the residual stream's width, so it can be passed through the model's final normalisation weights and output matrix to score every vocabulary token \citep[the ``logit lens'',][]{nostalgebraist2020}. We project the safety direction in raw coordinates at the fixed layer. In qwen-14b the highest-scoring tokens towards ``unsafe'' include ``anonymous'', ``sex'', ``rape'' and ``contraception'' for the base model and ``anonymous'', ``privacy'' and ``Sext'' for the distilled model; in qwen-32b, ``dispose'' and ``Anonymous'' for the base and ``Anonymous'' and ``desire'' for the distilled model. In llama-8b and math-7b the lists are word fragments with no readable content, although control directions decode to their own content in the same models. In the two families where the tokens are whole words, they are words that occur in unsafe prompts, both before and after distillation, though the specific words differ; no refusal words (``sorry'', ``cannot'') appear. This is an indirect reading and carries no weight in our conclusions.

\clearpage
\setcounter{figure}{0}\setcounter{table}{0}
\section{Preliminary run with reinforcement-learning checkpoints}
\label{app:earlier}
An earlier run of this project, whose pipeline is less fully recorded than the one reported here, also included for each family a checkpoint trained from the base by reinforcement learning alone, from SimpleRL-Zoo \citep{zeng2025}: hkust-nlp/Llama-3.1-8B-SimpleRL-Zoo, hkust-nlp/Qwen-2.5-14B-SimpleRL-Zoo and hkust-nlp/Qwen-2.5-32B-SimpleRL-Zoo, the only SimpleRL release for each of those bases; for Qwen2.5-Math-7B three SimpleRL releases exist and the one used was not recorded. Their activations on XSTest were almost identical to the base models' (mean per-prompt cosine 0.9998 or higher; relative distance 0.006 to 0.018), the base probe's AUC changed by at most 0.002, and the safety-direction cosine was at least 0.999. In the same run the distilled models' activations had mean per-prompt cosines of 0.79 to 0.96 with the base models'. A model that has barely changed cannot test whether a probe survives change, so these checkpoints say nothing about reinforcement learning in general and are not used in this paper.

\end{document}